\documentclass[sigconf]{acmart}

\AtBeginDocument{%
  \providecommand\BibTeX{{%
    \normalfont B\kern-0.5em{\scshape i\kern-0.25em b}\kern-0.8em\TeX}}}

\copyrightyear{2022}
\acmYear{2022}
\setcopyright{acmlicensed}\acmConference[MM '22]{Proceedings of the 30th ACM International Conference on Multimedia}{October 10--14, 2022}{Lisboa, Portugal}
\acmBooktitle{Proceedings of the 30th ACM International Conference on Multimedia (MM '22), October 10--14, 2022, Lisboa, Portugal}

\usepackage{booktabs}
\usepackage{multicol}
\usepackage{multirow}
\usepackage{graphicx}
\usepackage{subcaption}
\usepackage{array}
\usepackage{pifont}

\begin{document}

\title{ERNIE-mmLayout: Multi-grained MultiModal Transformer for Document Understanding}


\author{Wenjin Wang}
\authornote{Work done during internship at Baidu Inc..}
\email{wangwenjin@zju.edu.cn}
\orcid{0000-0001-7150-6162}
\affiliation{
  \institution{Zhejiang University}
  \city{Hangzhou}
  \country{China}
}

\author{Zhengjie Huang}
\email{huangzhengjie@baidu.com}
\orcid{0000-0003-1878-0554}
\affiliation{
  \institution{Baidu Inc.}
  \country{China}
}

\author{Bin Luo}
\email{luobin06@baidu.com}
\affiliation{
  \institution{Baidu Inc.}
  \country{China}
}

\author{Qianglong Chen}
\email{chenqianglong@zju.edu.cn}
\affiliation{
  \institution{Zhejiang University}
  \city{Hangzhou}
  \country{China}
}

\author{Qiming Peng, Yinxu Pan}
\email{{pengqiming, panyinxu}@baidu.com}
\affiliation{
  \institution{Baidu Inc.}
  \country{China}
}

\author{Weichong Yin, Shikun Feng}
\email{{yinweichong, fengshikun01}@baidu.com}
\affiliation{
  \institution{Baidu Inc.}
  \country{China}
}

\author{Yu Sun, Dianhai Yu}
\email{{sunyu02, yudianhai}@baidu.com}
\affiliation{
  \institution{Baidu Inc.}
  \country{China}
}

\author{Yin Zhang}
\authornote{Corresponding author: Yin Zhang.}
\email{zhangyin98@zju.edu.cn}
\affiliation{
  \institution{Zhejiang University}
  \city{Hangzhou}
  \country{China}
}



\renewcommand{\shortauthors}{Wenjin Wang et al.}

\begin{abstract}
Recent efforts of multimodal Transformers have improved Visually Rich Document Understanding (VrDU) tasks via incorporating visual and textual information.
However, existing approaches mainly focus on fine-grained elements such as words and document image patches, making it hard for them to learn from coarse-grained elements, including natural lexical units like phrases and salient visual regions like prominent image regions.
In this paper, we attach more importance to coarse-grained elements containing high-density information and consistent semantics, which are valuable for document understanding.
At first, a document graph is proposed to model complex relationships among multi-grained multimodal elements, in which salient visual regions are detected by a cluster-based method. 
Then, a \textbf{m}ulti-grained \textbf{m}ultimodal Transformer called \textbf{mmLayout} is proposed to incorporate coarse-grained information into existing pre-trained fine-grained multimodal Transformers based on the graph.
In \textbf{mmLayout}, coarse-grained information is aggregated from fine-grained, and then, after further processing, is fused back into fine-grained for final prediction.
Furthermore, common sense enhancement is introduced to exploit the semantic information of natural lexical units.
Experimental results on four tasks, including information extraction and document question answering, show that our method can improve the performance of multimodal Transformers based on fine-grained elements and achieve better performance with fewer parameters.
Qualitative analyses show that our method can capture consistent semantics in coarse-grained elements.
\end{abstract}



\maketitle

\section{Introduction}
\begin{figure}[ht]
  \small
  \centering
  \includegraphics[width=1.0\columnwidth,trim={0.0cm 0.30cm 0.0cm 0.0cm},clip]{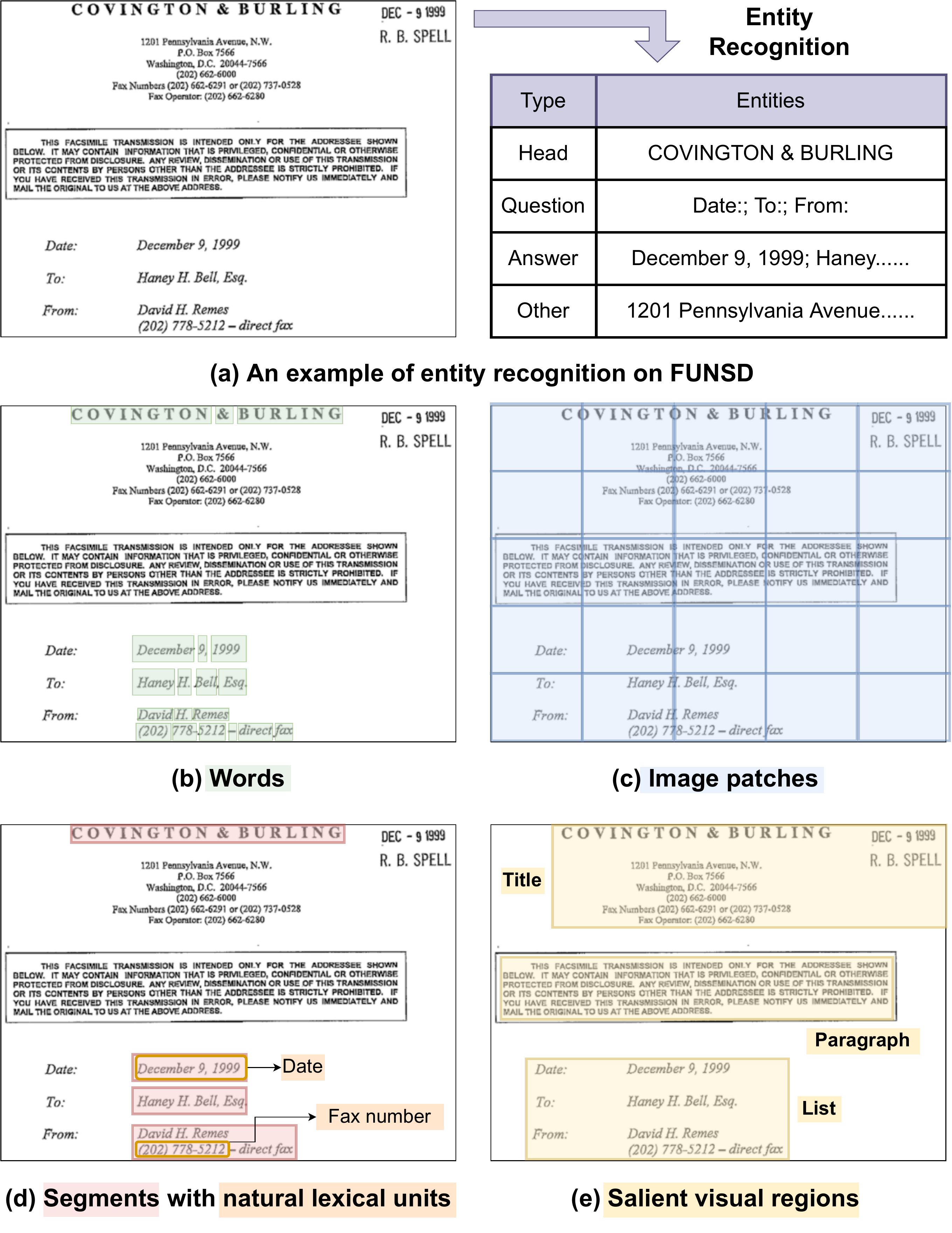}
\setlength{\abovecaptionskip}{0.1cm}
\setlength{\belowcaptionskip}{0.1cm}
   \caption{(a) Entity recognition on FUNSD. (b,c) Fine-grained elements words and image patches. (d, e) Coarse-grained elements natural lexical units and salient visual regions.}
  \label{fig:motivation}
\end{figure}

Visually-rich Document Understanding (VrDU) is a critical component of document intelligence \cite{cuiDocumentAIBenchmarks2021} that aims to understand scanned or digital-born documents.
Despite many advances in vision-language understanding, extracting structural information in visually-rich documents remains a major challenge because it involves different types of information, including image, text, and layout.

Extensive efforts have been made to solve this challenge based on CNNs \cite{yangLearningExtractSemantic2017,kattiChargridUnderstanding2D2018,denkBERTgridContextualizedEmbedding2019,zhaoCUTIELearningUnderstand2019,sarkhelDeterministicRoutingLayout2019,zhangTRIEEndtoEndText2020,wangRobustVisualInformation2021,linViBERTgridJointlyTrained2021} and GNNs \cite{liuGraphConvolutionMultimodal2019,qianGraphIEGraphBasedFramework2019,yuPICKProcessingKey2020,weiRobustLayoutawareIE2020,carbonellNamedEntityRecognition2021}.
Recently, many pre-trained multimodal Transformers for VrDU \cite{xuLayoutLMPretrainingText2020,appalarajuDocFormerEndtoEndTransformer2021a,garncarekLAMBERTLayoutAwareLanguage2021a,hwangSpatialDependencyParsing2021,liStructuralLMStructuralPretraining2021,liSelfDocSelfSupervisedDocument2021,liStrucTexT2021,powalskiGoingFullTILTBoogie2021,xuLayoutLMv2MultimodalPretraining2021a,xuLayoutXLMMultimodalPretraining2021,hongBROSPreTrainedLanguage2022,leeFormNetStructuralEncoding2022} have been proposed, inspired by the success of pre-training in vision language understanding.
They incorporate layout information into models by various 2D position embedding and spatial-aware self attention mechanisms. Combined with carefully designed pre-training tasks, they make amazing progress on many tasks.

However, existing layout-aware multimodal Transformers mainly focus on fine-grained information such as words and image patches.
They ignore coarse-grained information including natural lexical units \cite{zhangAMBERTPretrainedLanguage2021,guoLICHEEImprovingLanguage2021}, like multi-word expressions and phrases, and salient visual regions \cite{andersonBottomUpTopDownAttention2018,leeStackedCrossAttention2018}, like attractive or prominent image regions.
These coarse-grained elements contain high-density information and consistent semantics, which are valuable for document understanding.
For example, as shown in Figure~\ref{fig:motivation}(b) and Figure~\ref{fig:motivation}(d), given the tokens ``(202)'', ``778-'', and ``5212'' that belong to the fax number ``(202) 778-5212'', it is difficult to directly determine that a single token itself is a part of a fax number.
On the contrary, we will get this information more easily if we look at them as a whole.
The potential of such natural lexical units has been demonstrated in natural language models \cite{zhangAMBERTPretrainedLanguage2021,guoLICHEEImprovingLanguage2021}.
Similarly, as shown in Figure~\ref{fig:motivation}(c) and Figure~\ref{fig:motivation}(e), compared with image patches, salient visual regions can reflect richer semantic information.
If we can determine that a region corresponds to a list by visual information, combined with layout relationships, we can realize that it may contain multiple pairs of Question and Answer (in FUNSD).

Several efforts \cite{liStructuralLMStructuralPretraining2021,liStrucTexT2021,liSelfDocSelfSupervisedDocument2021} have attempted to take coarse-grained information into account, but the effects of coarse-grained information have not been thoroughly evaluated and coarse-grained visual information has been ignored.
StructuralLM \cite{liStructuralLMStructuralPretraining2021} only considers the coarse-grained layout information and replaces token level 2D layout embeddings with segment level text 2D layout embeddings.
Segment level text features in StrucText \cite{liStrucTexT2021} are directly aggregated from the token level features, without sufficient interaction with other information.
SelfDoc \cite{liSelfDocSelfSupervisedDocument2021} extracts features by the Sentence-BERT \cite{reimersSentenceBERT2019} and ignores word level embeddings.

Our study takes into account both fine-grained and coarse-grained multimodal information for visually rich document understanding.
We focus on how to effectively incorporate coarse-grained information into existing pre-trained Transformers with fine-grained information, rather than designing new pre-training tasks like previous attempts.
To this end, we regard a document as a graph with nodes corresponding to multi-grained multimodal elements in the document.
Based on the graph, we propose the multi-grained multimodal Transformer named \textbf{mmLayout} consists of four modules: two graph attention modules called  \emph{Fine-grained Encoder} and \emph{Coarse-grained Encoder}, and two hierarchical modules called \emph{Cross-grained Aggregation} and \emph{Cross-grained Fusion}.
The main challenges in our method are (1) the lack of natural salient visual regions in raw data or results of OCR tools and (2) the inconsistency between text segments obtained by OCR tools and natural lexical units.
For challenge (1), a clustering-based method is proposed to detect salient visual regions according to the distribution of textual segments.
The intuitive idea behind it is that the salience of a region in documents is related to the concentration of texts in it.
For challenge (2), we introduce the Common Sense Enhancement strategy to exploit natural lexical units in text segments by detecting common knowledge in them.
This knowledge is integrated into features of text segments to compensate for the loss of important semantic information.

Our contributions are summarized as follows:
\begin{itemize}
    \item{
    We consider the importance of coarse-grained elements containing high-density information and consistent semantics, which are valuable for document understanding.
    To model complex relations among various elements and incorporate coarse-grained information into existing multimodal Transformers, we construct a document graph and propose a multi-grained multimodal Transformer named mmLayout.}
    \item{We propose a clustering-based method to detect salient visual regions and introduce a Common Sense Enhancement strategy to exploit the semantic information of natural lexical units in text segments.}
    \item{Experimental results on information extraction (FUNSD, CORD, SROIE) and document question answering (DocVQA) tasks show that our method significantly performs better than multimodal Transformers based on fine-grained elements and achieves better performance with fewer parameters.
    Qualitative analyses show that our method can better capture consistent semantics in coarse-grained elements.}
\end{itemize}

\section{Preliminaries}
\label{sec:pre}
In this section, we briefly describe the key components in spatial-aware multimodal Transformer and refer to Section~\ref{sec:related_work} and LayoutLMv2 \cite{xuLayoutLMv2MultimodalPretraining2021a} for more details.

\textbf{Input representation.} 
The input representation of a document page image $\mathcal{D}$ consists of textual, visual, and layout embeddings. Given a document page, OCR tools are adopted to obtain a sequence of textual tokens $\mathcal{W}=\{w_1,w_2,\dots,w_L\}$ and corresponding bounding boxes $\mathcal{B}_t=\{b_1,b_2,\dots,b_L\}$ where $L$ represents the length of the sequence (Fig.~\ref{fig:motivation}(b))\cite{xuLayoutLMPretrainingText2020,xuLayoutLMv2MultimodalPretraining2021a,liStructuralLMStructuralPretraining2021}.
A bounding box $b=(x^0,y^0,x^1,y^1)$ where $(x^0,y^0)$ and $(x^1,y^1)$ are coordinates of top left and bottom right corners of $w$.
Then, the document image $\mathcal{D}$ is processed by a visual encoder based on the ResNeXt-FPN architecture \cite{xieAggregatedResidualTransformations2017,linFeaturePyramidNetworks2017}.
The output feature map is pooled, flattened, and projected into a sequence of visual token features denoted as $\mathbf{I}=\{\mathbf{I}_1, \mathbf{I}_2, \dots, \mathbf{I}_{W\times H}\}$ where $W\times H$ is the number of visual tokens \cite{xuLayoutLMv2MultimodalPretraining2021a}. The corresponding bounding boxes are denoted as $\mathcal{B}_v=\{b_1,b_2,\dots,b_{W\times H}\}$ which are obtained by dividing the document images into $W\times H$ patches.

The textual embedding layer \cite{xuLayoutLMv2MultimodalPretraining2021a} is constructed as follows:
\begin{equation}
    \label{eq:textual_emb}
    E_\text{text}(\mathcal{W})=\{t_i\mid t_i=E_\text{w}(w_i)+E_\text{t}(sent_i)+E_\text{p}(i),i\in[1,L]\},
\end{equation}
where $E_\text{w}$, $E_\text{t}$, and $E_\text{p}$ represent the word token, token type, and 1D position embedding layer respectively.

Similarly, the visual embedding layer \cite{xuLayoutLMv2MultimodalPretraining2021a} is constructed as follows:
\begin{equation}
    \label{eq:visual_emb}
    E_\text{visual}(\mathbf{I})=\{v_i\mid v_i=\mathbf{I}_i+E_\text{t}(sent_i)+E_\text{p}(i),i\in[1,W\times H]\}.
\end{equation}
Note that $E_\text{t}$ and $E_\text{p}$ are shared by textual and visual embedding layers.

The layout embedding layer \cite{xuLayoutLMv2MultimodalPretraining2021a} is constructed as follows:
\begin{equation}
\begin{aligned}
    \label{eq:layout_emb}
    E_\text{layout}(\mathcal{B})=\{&\text{Concat}(E_X(x_i^0),E_X(x_i^1),E_X(x_i^1-x_i^0),\\
    &E_Y(y_i^0),E_Y(y_i^1),E_Y(y_i^1-y_i^0)),i\in[1,L']\},
\end{aligned}
\end{equation}
where $\mathcal{B}$ represents a sequence of bounding boxes with length $L'$, and $E_X$ and $E_Y$ are x-axis and y-axis embedding layers respectively.

At last, the input representation is obtained as follows:
\begin{equation}
    \mathcal{I}=\text{Concat}(E_\text{text}(\mathcal{W}),E_\text{visual}(\mathbf{I}))+E_\text{layout}(\mathcal{B}_t\cup\mathcal{B}_v).
\end{equation}

\textbf{Spatial-Aware Self-Attention.} 
In self-attention, the inputs $H\in\mathbb{R}^{n\times d}$ are linearly transformed to queries $Q\in\mathbb{R}^{n\times d_k}$, keys $K\in\mathbb{R}^{n\times d_k}$, and values $V\in\mathbb{R}^{n\times d_v}$, where $n$ is the input sequence length, $d$, $d_k$, $d_v$ are the dimensions of inputs, queries (keys) and values, respectively. The canonical self-attention is written as follows:
\begin{equation}
    \text{softmax}(\frac{QK^T}{\sqrt{d_k}})V=AV,
\end{equation}
where $A$ is the attention matrix. To explicitly introduce layout information into canonical self-attention mechanism, learnable relative 1D and 2D position bias is added to original attention matrix as follows:
\begin{equation}
    A_{ij}'=A_{ij}+\mathbf{R}_\text{1D}(j-i)+\mathbf{R}_X(x_j^0-x_i^0)+\mathbf{R}_Y(y_j^0-y_i^0),
\end{equation}
where $A'$ is the spatial-aware attention matrix, $\textbf{R}_\text{1D}$, $\textbf{R}_X$, $\textbf{R}_Y$ are relative 1D and 2D position embedding layers, respectively. For convenience, in this paper, we represent the spatial-aware multi-head self-attention as follows:
\begin{equation}
\label{eq:spatial_MHA}
    \textbf{Spatial-MHA}(H,\mathcal{B})=A'V,
\end{equation}
where $H$ is the input features and $\mathcal{B}$ is the layout information of bounding boxes.

\section{Method}
\label{sec:method}
\begin{figure*}[t]
\small
  \centering
  \includegraphics[width=0.98\linewidth]{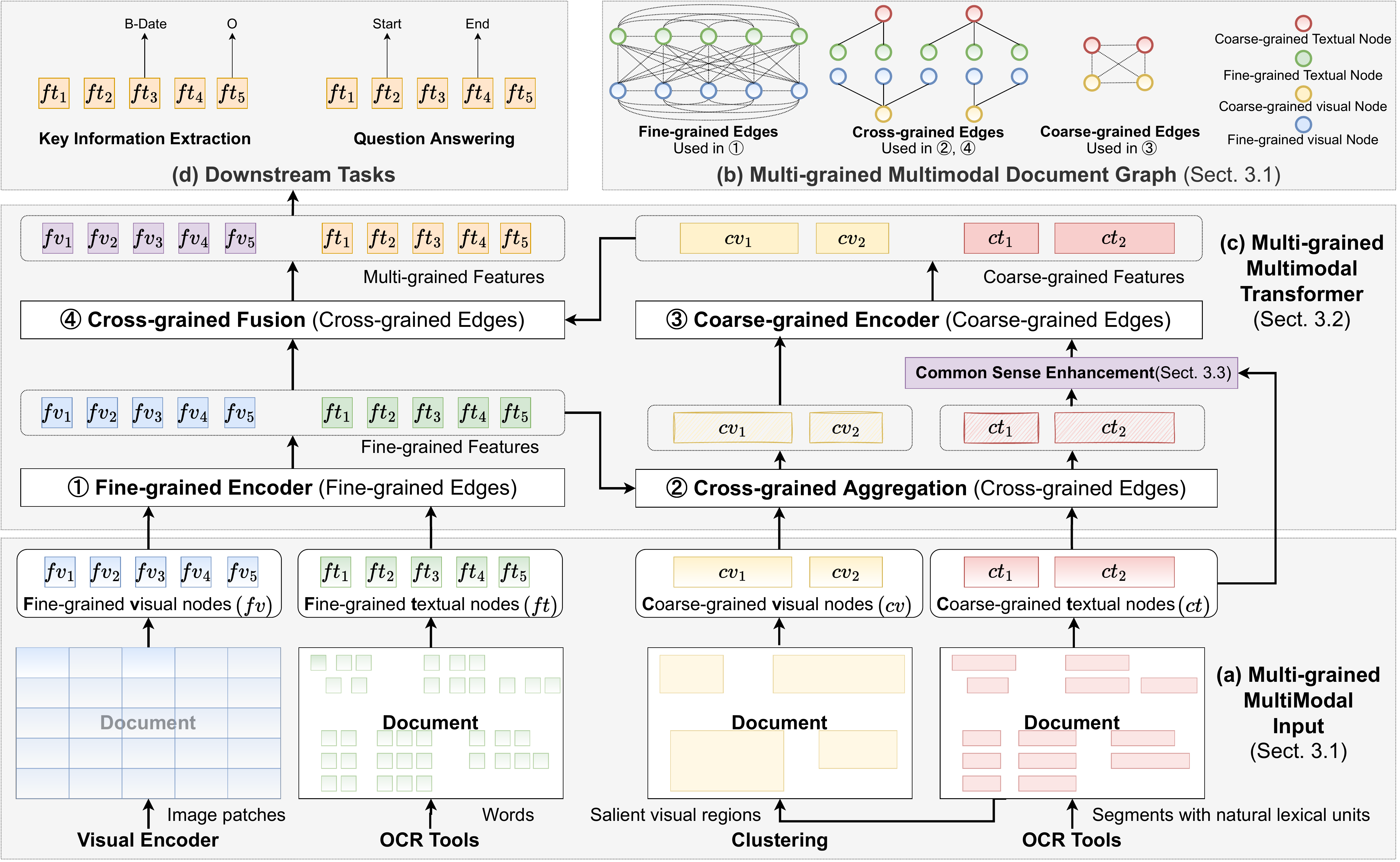}
  
\setlength{\abovecaptionskip}{0.15cm}
  \caption{\textbf{Overall architecture of mmLayout.} (a) Multi-grained Multimodal Input: a clustering-based method is proposed to construct coarse-grained visual nodes. (b) Multi-grained Multimodal Document Graph: it models complex relations among multi-grained multimodal elements in document.
  (c) Multi-grained Multimodal Transformer (mmLayout): \ding{172} Fine-grained semantic features are extracted by the Fine-grained Encoder
  \ding{173} Fine-grained features are aggregated according to cross-grained edges and common sense enhancement is applied (in Fig.~\ref{fig:common_sense}). \ding{174} Coarse-grained semantic features are extracted by the Coarse-grained Encoder. \ding{175} Fine-grained and coarse-grained features are fused to generate multi-grained features for prediction.}
  \Description{}
  \label{fig:architecture}
\end{figure*}

We propose \emph{mmLayout}, a multi-grained multimodal Transformer, to incorporate coarse-grained information into existing layout-aware multimodal pre-trained Transformers and the illustration is shown in Figure~\ref{fig:architecture}.
At first, a document graph is proposed whose nodes are multi-grained multimodal document elements and edges are used to model complex relationships among these elements, including fine-grained, coarse-grained, and cross-grained edges (see Fig.~\ref{fig:architecture}(a) and Fig.~\ref{fig:architecture}(b)).
Then, the \emph{mmLayout} extracts semantic features by four modules: two graph attention modules called \emph{Fine-grained Encoder} and \emph{Coarse-grained Encoder}, and two hierarchical modules called \emph{Cross-grained Aggregation} and \emph{Cross-grained Fusion}, based on the graph (see Fig.~\ref{fig:architecture}(c)).
Recent works have shown that Transformers are competitive graph encoders \cite{caiGraphTransformerGraphtoSequence2019,yingTransformersReallyPerform2021}, therefore we adopt the Transformer architecture for the above-mentioned encoders to model interaction along fine-grained and coarse-grained edges in the document graph.
Guided by cross-grained edges, fine-grained information is aggregated into coarse-grained, and then the output of the \emph{Coarse-grained Encoder} is fused back into fine-grained for final prediction.
Moreover, we propose a common sense enhancement strategy to avoid semantic confusion in features of textual segments (Sec~\ref{sec:common}).

\subsection{Document Graph}
\label{sec:document_graph}
Given a document page image $\mathcal{D}$, the corresponding document graph is defined as $\mathcal{G}=(\mathcal{V},\mathcal{E})$ where $\mathcal{V}$ and $\mathcal{E}$ represent nodes and edges respectively. The document graph contains four types of nodes: fine-grained textual, fine-grained visual, coarse-grained textual, and coarse-grained visual nodes, which are denoted as $\mathcal{V}_{ft}$, $\mathcal{V}_{fv}$, $\mathcal{V}_{ct}$, and $\mathcal{V}_{cv}$, respectively. And it contains three types of edges: fine-grained, coarse-grained, and cross-grained edges denoted as $\mathcal{E}_{f}$, $\mathcal{E}_{c}$ and $\mathcal{E}_{cf}$ respectively.

\textbf{Fine-grained nodes.} We regard words and image patches described in Section~\ref{sec:pre} as fine-grained textual nodes $\mathcal{V}_{ft}=\mathcal{W}$ and visual nodes $\mathcal{V}_{fv}=\mathcal{I}$ respectively. The corresponding bounding boxes are denoted as $\mathcal{B}_{ft}=\mathcal{B}_{t}$ and $\mathcal{B}_{fv}=\mathcal{B}_{v}$.

\textbf{Coarse-grained textual nodes based on segments.}
A sequence $\mathcal{S}=\{s_1,s_2\dots,s_Z\}$ with $Z$ text segments obtained from the document image by OCR tools are regarded as coarse-grained textual nodes $\mathcal{V}_{ct}=\mathcal{S}$.
The corresponding bounding boxes are denoted as $\mathcal{B}_{ct}=\{b^{ct}_1,b^{ct}_2,\dots,b^{ct}_Z\}$.

\textbf{Coarse-grained visual nodes based on clustering.}
We regard salient visual regions as coarse-grained visual nodes.
Due to the lack of natural salient visual regions, we detect them by applying a density-based clustering method DBSCAN \cite{ester1996DBSCAN,khanDBSCANPresentFuture2014} to coarse-grained textual nodes.
The intuitive idea is that the salience of a region is related to the concentration of texts in it.
Specifically, we views textual nodes clusters as high-density regions separated by low-density regions.
A node is in regions of high-density if there exist $m$ other nodes with a distance of $r$ where $m$ and $r$ are hyper-parameters.
We define the distance between nodes $\mathcal{V}_{ct}^i$ and $\mathcal{V}_{ct}^j$ as follows:
\begin{align}
    \text{dist}(i,j)&=\sqrt{(\text{dist}_x(i,j))^2+(\text{dist}_y(i,j))^2}, \\
    \text{dist}_x(i,j)&=\max(\max(x_i^0,x_j^0)-\min(x_i^1,x_j^1),0), \\
    \text{dist}_y(i,j)&=\max(\max(y_i^0,y_j^0)-\min(y_i^1,y_j^1),0),
\end{align}
where $\text{dist}_x(i,j)$ and $\text{dist}_y(i,j)$ represent the horizontal and vertical distance between the boundaries of two boxes.
Some clusters of coarse-grained text nodes are obtained and regarded as coarse-grained visual nodes.
The bounding box of a coarse-grained visual node is a rectangle that just covers all bounding boxes of coarse-grained textual nodes in the corresponding cluster.
We denote the coarse-grained visual nodes and their bounding boxes as $\mathcal{V}_{cv}$ and $\mathbf{B}_{cv}=\{b_1^{cv},b_2^{cv},\dots,b_P^{cv}\}$ respectively where $P$ is the number of coarse-grained visual nodes. One reason to choose DBSCAN is that it can generate different numbers of clusters for different documents, accommodating diversity of document layouts.
The number of salient visual regions is affected by the radius $r$ (see Sect~\ref{sec:exp_ablation}). 

\textbf{Edges.}
Three types of edges are constructed (see Fig.~\ref{fig:architecture}(b)).
It is difficult to directly model the fine-grained edges $\mathcal{E}_f$ and coarse-grained edges $\mathcal{E}_c$ among multimodal nodes. So we regard them as fully connected and learn soft connections by global graph attention, which is equivalent to self-attention in Transformer (see Fig.~\ref{fig:architecture}(c)).
Instead, cross-grained edges $\mathcal{E}_{cf}$ among multi-grained single modal nodes are directly determined by the layout.
Specifically, a fine-grained textual node, i.e. a word, is connected to a coarse-grained textual node, i.e. a text segment, containing it.
A fine-grained visual node is connected to the coarse-grained visual node with the largest IOU(Intersection over Union) of its bounding box.
Note that a coarse-grained node can be connected to multiple fine-grained nodes of the same modal, whereas a fine-grained node can only be connected to one coarse-grained node.

\subsection{Model Architecture}
\label{sec:model}
Given a document graph $\mathcal{G}$, fine-grained semantic features are extracted by the \emph{Fine-grained Encoder}. Then, coarse-grained semantic features are extracted by the \emph{Cross-grained Aggregation} and \emph{Coarse-grained Encoder}. At last, coarse-grained features are fused back into fine-grained by the \emph{Cross-grained Fusion} for prediction.

\textbf{Input representation.}
Given the document graph $\mathcal{G}=(\mathcal{V},\mathcal{E})$, the input embedding for the \emph{Fine-grained Encoder} is as follows:
\begin{equation}
    \mathcal{I}_f=\text{Concat}(E_\text{text}(\mathcal{V}_{ft})+E_\text{layout}(\mathcal{B}_{ft}),E_\text{visual}(\mathcal{V}_{fv})+E_\text{layout}(\mathcal{B}_{fv})),
\end{equation}
where $E_\text{text}$, $E_\text{visual}$, and $E_\text{layout}$ are the textual, visual, and layout embedding layers defined in Equation~(\ref{eq:textual_emb}), (\ref{eq:visual_emb}), and (\ref{eq:layout_emb}) in Section~\ref{sec:pre} respectively.

\textbf{Fine-grained Encoder.}
The \emph{Fine-grained Encoder} takes fine-grained spatial-aware multimodal embedding as input and generates context-aware features for fine-grained nodes by graph attention according to fully connected fine-grained edges (see Fig.~\ref{fig:architecture}(b)). It consists of a stack of spatial-aware multimodal Transformer layers to learn interaction between fine-grained elements as follows:
\begin{align}
    \mathbf{H}_{f,i+1}=\text{LN}(\text{FFN}(\text{LN}(\text{Spatial-MHA}(\mathbf{H}_{f,i},\{\mathcal{B}_{ft},\mathcal{B}_{fv}\})))),
\end{align}
where $\text{Spatial-MHA}$ is the spatial-aware self-attention (see Sect.~\ref{sec:pre}), and $\text{LN}$ and $\text{FFN}$ are the layer norm and fully connected feed-forward network respectively.
The $\mathbf{H}_{f,i}$ is the output of layer $i$ and the initial input is set to $\mathbf{H}_{f,0}=\mathcal{I}_f$. 
The output of the \emph{Fine-grained Encoder} is denoted to $\mathbf{H}_{f,N}$ where $N$ is the number of layers.
The \emph{Fine-grained Encoder in our model is consistent with many existing pre-trained layout-aware multimodal Transformers, allowing us to directly leverage their capabilities.}
And it can be also viewed as performing graph attention between fine-grained nodes in the document graph.
To incorporate coarse-grained information into $\mathbf{H}_{f,N}$, we further process it with the \emph{Coarse-grained Encoder}. 

\textbf{Coarse-grained Aggregation.}
At first, the initial features of coarse-grained nodes $\mathbf{I}_{c}'$ are obtained by aggregating fine-grained features $\mathbf{H}_{f,N}$ according to cross-grained edges $\mathcal{E}_{cf}$ as follows:
\begin{equation}
    \mathbf{I}_{c}'=\{h_{c,0}^{'i}\mid h_{c,0}^{'i}=\sum_{(i,j)\in\mathcal{E}_{cf}}\mathbf{H}_{f,N}^{j}\}=\{\mathcal{I}_{c,t}',\mathcal{I}_{c,v}'\},
    \label{eq:aggregation}
\end{equation}
where $\mathcal{I}_{c,t}'$ and $\mathcal{I}_{c,v}'$ represent the features of coarse-grained textual and visual nodes respectively.
To enhance the semantic information of coarse-grained textual nodes, the common knowledge extracted from text segments is fused into features $\mathcal{I}_{c,t}'$, and the new features are denoted as $\mathcal{I}_{c,t}''$ (details in Sect.~\ref{sec:common}).
Combined with the coarse-grained layout information, the coarse-grained input representation is as follows:
\begin{equation}
    \mathcal{I}_c=\text{Concat}(\mathcal{I}_{c,t}''+E_\text{layout}(\mathcal{B}_{ct}),\mathcal{I}_{c,v}'+E_\text{layout}(\mathcal{B}_{cv})),
\end{equation}
where $E_\text{layout}$ represents the layout embedding that is shared with the \emph{Fine-grained Encoder}.

\textbf{Coarse-grained Encoder.}  To extract the relations and semantics from coarse-grained elements connected by coarse-grained edges (see Fig.~\ref{fig:architecture}(b)), we apply another stack of canonical Transformer layers as \emph{Coarse-grained Encoder} as follows:
\begin{equation}
    \mathbf{H}_{c,i+1}=\text{LN}(\text{FFN}(\text{LN}(\text{MHA}(\mathbf{H}_{c,i})))),
\end{equation}
where $\text{MHA}$ is the canonical multi-head self-attention mechanism.
The $\mathbf{H}_{c,i}$ is the output of layer $i$ and the initial input is set to $\mathbf{H}_{c,0}=\mathcal{I}_c$. The output of the \emph{Coarse-grained Encoder} is denoted to $\mathbf{H}_{c,M}$ where $M$ is the number of layers.
We also try to use spatial-aware self-attention in \emph{Coarse-grained Encoder}, but find no additional benefits compared to the canonical version.

Experimental results show that equipping existing base models with a few layers of \emph{Coarse-grained Encoder} can outperform existing large models on multiple datasets. Although the \emph{Coarse-grained Encoder} introduces additional parameters on base models, it keeps a smaller total number of parameters than the large one.

\textbf{Cross-grained Fusion.} Finally, for each fine-grained node, we fuse the feature of itself and the feature of the coarse-grained node corresponding to it as follows:
\begin{equation}
    \mathbf{H}=\{h^i\mid h^i=\mathbf{H}_{f,N}^i+\mathbf{H}_{c,M}^j,(i,j)\in\mathcal{E}_{cf}\},  
\end{equation}
where $i\in(1,L+W\times H)$ and $\mathbf{H}$ is used for different tasks.

\subsection{Common Sense Enhancement}
\label{sec:common}
\begin{figure}[t]
  \centering
  \small
  \includegraphics[width=1.0\columnwidth]{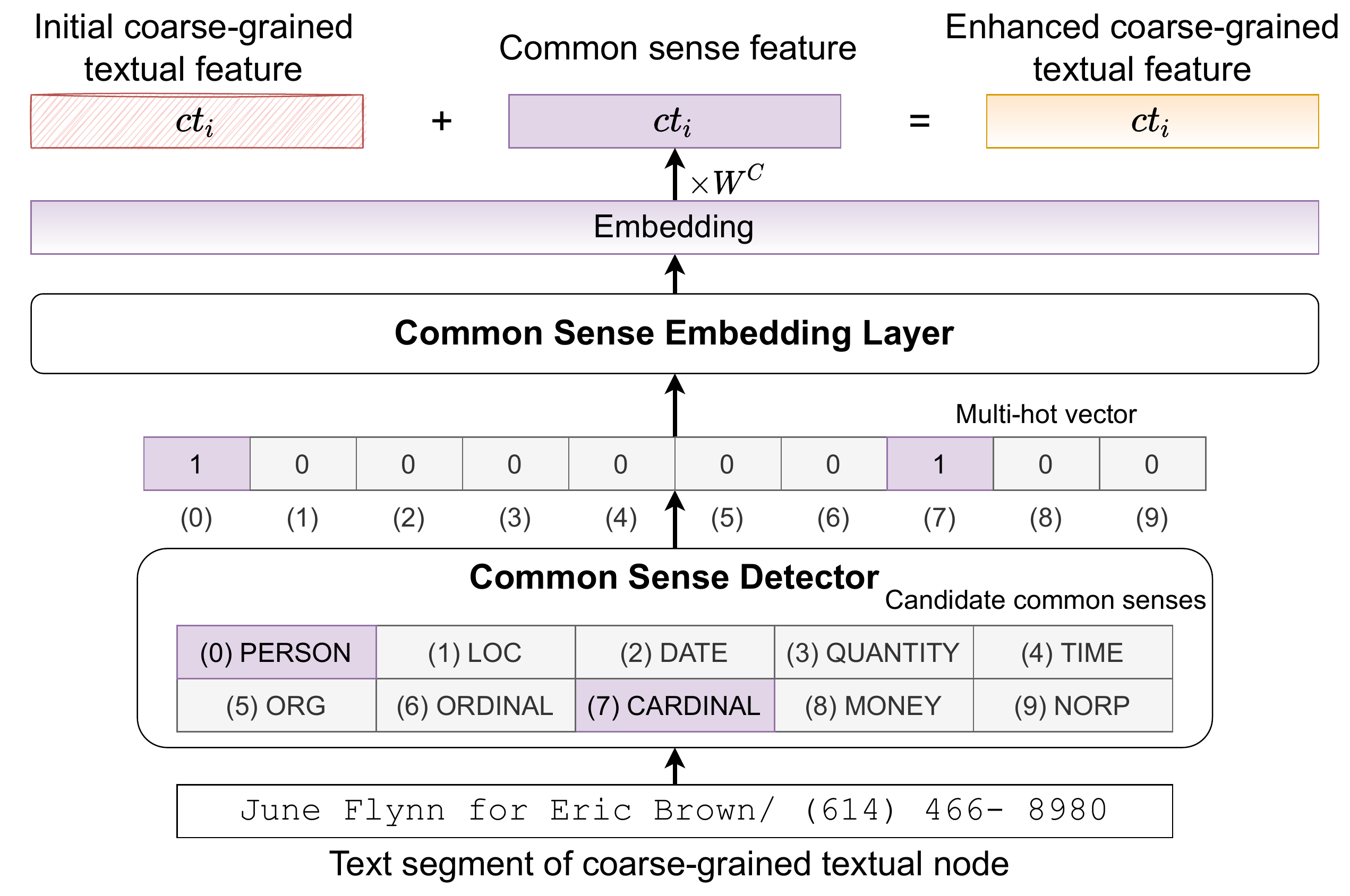}
\setlength{\abovecaptionskip}{0.15cm}
  \caption{Common sense enhancement. The Common Sense Detector detects common senses contained in the text segment from candidate common senses and generates a multi-hot vector to obtain the final common sense embedding.}
  \Description{.}
  \label{fig:common_sense}
\end{figure}
To obtain feature of a coarse-grained textual node, features of fine-grained textual nodes belonging to it are directly aggregated (in Equation~(\ref{eq:aggregation})).
However, simple aggregation may result in the semantic confusion caused by the fact that a textual segment may contain multiple natural lexical units.
To solve this problem, we propose a common sense enhancement strategy to detect common knowledge in natural semantic units contained in segments.
Then, we incorporate this knowledge into features of segments to compensate for the loss of important semantic information.
Specifically, given a coarse-grained textual node $\mathcal{V}_{ct}^i$ with initial feature $\mathcal{I}_{c,t}'^i$ and raw text segment $s_i$, we detect common senses contained in the text segment from $K$ candidate common senses by an open source tool spaCy \footnote{https://spacy.io/} (see Fig.~\ref{fig:common_sense}) and generate a multi-hot vector $\mathcal{C}_i\in\mathbb{R}^{K}$. Then, we create a learnable common sense embedding matrix $E_\text{c}$ with size $K\times d$ and $E_\text{c}^k$ represents the embedding of common sense entity type $k$. The common sense embedding for node $\mathcal{V}_{ct}^i$ is calculated as follows:
\begin{equation}
    \mathcal{C}'_i=W^CE_c^T\mathcal{C}_i,
\end{equation}
where $W^C$ is a projection transform for dimensional alignment and $\mathcal{I}_{c,t}''^i=\mathcal{I}_{c,t}'^i+\mathcal{C}'_i$. 
Note that common sense entity categories are different from the categories in the downstream named entity recognition tasks and are task-independent. The core idea is that common sense information, such as person and institution, is helpful for model to understand the document.

\section{Experiments}
\begin{table*}[htbp]
\setlength{\abovecaptionskip}{0.15cm}
\setlength{\belowcaptionskip}{0.01cm}
\caption{Performance (entity level F1 score) on FUNSD, CORD, and SROIE.}
\label{tab:exp_funsd_cord_sroie}
\begin{tabular}{l>{\centering\arraybackslash}p{1cm}>{\centering\arraybackslash}p{1cm}>{\centering\arraybackslash}p{1cm}>{\centering\arraybackslash}p{1cm}>{\centering\arraybackslash}p{1cm}>{\centering\arraybackslash}p{1cm}ccc}
    \toprule
    \multirow{2}{*}{Method} &
    \multicolumn{3}{c}{Fine-grained Information} &
     \multicolumn{3}{c}{Coarse-grained Information} &
    \multirow{2}{*}{FUNSD} &
    \multirow{2}{*}{CORD} &
    \multirow{2}{*}{SROIE} \\
    &
    Text &
    Vision &
    Layout &
    Text &
    Vision &
    Layout &
    &
    &
    \\
    \midrule
    BERT$_\text{base}$ &
    \checkmark &
    &
    &
    &
    &
    &
    0.6026 &
    0.8968 &
    0.9099 \\
    UniLMv2$_\text{base}$ &
    \checkmark &
    &
    &
    &
    &
    &
    0.6648 &
    0.9092 &
    0.9459 \\
    \midrule
    BROS$_\text{base}$ &
    \checkmark &
    &
    \checkmark &
    &
    &
    &
    0.8305 &
    0.9650 &
    0.9628 \\
    \midrule
    LayoutLM$_\text{base}$ &
    \checkmark &
    \checkmark &
    \checkmark &
    &
    &
    &
    0.7866 &
    0.9472 &
    0.9438 \\
    LayoutLMv2$_\text{base}$ &
    \checkmark &
    \checkmark &
    \checkmark &
    &
    &
    &
    0.8276 &
    0.9495 &
    0.9625 \\
    DocFormer$_\text{base}$ &
    \checkmark &
    \checkmark &
    \checkmark &
    &
    &
    &
    0.8334 &
    0.9633 &
    - \\
    \midrule
    StrucText$_\text{base}$ &
    \checkmark &
    \checkmark &
    \checkmark &
    \checkmark &
    &
    &
    0.8309 &
    - &
    0.9688 \\
    SelfDoc$_\text{base}$ &
    \checkmark &
    \checkmark &
    \checkmark &
    \checkmark &
    &
    &
    0.8336 &
    - &
    - \\
    \midrule
    mmLayout$_\text{base}$ &
    \checkmark &
    \checkmark &
    \checkmark &
    \checkmark &
    \checkmark &
    \checkmark &
    \textbf{0.8602} &
    \textbf{0.9723} &
    \textbf{0.9763} \\
    \bottomrule
    BERT$_\text{large}$ &
    \checkmark &
    &
    &
    &
    &
    &
    0.6563 &
    0.9025 &
    0.9200 \\
    UniLMv2$_\text{large}$ &
    \checkmark &
    &
    &
    &
    &
    &
    0.7072 &
    0.9205 &
    0.9488 \\
    \midrule
    BROS$_\text{large}$ &
    \checkmark &
    &
    \checkmark &
    &
    &
    &
    0.8452 &
    0.9728 &
    0.9662 \\
    \midrule
    LayoutLM$_\text{large}$ &
    \checkmark &
    \checkmark &
    \checkmark &
    &
    &
    &
    0.7895 &
    0.9493 &
    0.9524 \\
    LayoutLMv2$_\text{large}$ &
    \checkmark &
    \checkmark &
    \checkmark &
    &
    &
    &
    0.8420 &
    0.9601 &
    0.9781 \\
    DocFormer$_\text{large}$ &
    \checkmark &
    \checkmark &
    \checkmark &
    &
    &
    &
    0.8455 &
    0.9699 &
    - \\
    \midrule
    StructuralLM$_\text{large}$ &
    \checkmark &
    \checkmark &
    \checkmark &
    &
    &
    \checkmark &
    0.8514 &
    - &
    - \\
    \midrule
    mmLayout$_\text{large}$ &
    \checkmark &
    \checkmark &
    \checkmark &
    \checkmark &
    \checkmark &
    \checkmark &
    \textbf{0.8649} &
    \textbf{0.9738} &
    \textbf{0.9791} \\
    \bottomrule
\end{tabular}
\end{table*}

\subsection{Experimental Settings}
\textbf{Datasets.} We conduct experiments on three Information Extraction tasks, including FUNSD \cite{jaumeFUNSDDatasetForm2019}, CORD \cite{parkCORDConsolidatedReceipt2019}, and SROIE \cite{huangICDAR2019Competition2019}, and a Document Question Answering task DocVQA \cite{mathewDocVQADatasetVQA2021}.
We regard the entity extraction on FUNSD, CORD and SROIE as sequential labeling tasks and evaluate performance on them with entity-level F1 score. We evaluate the performance on DocVQA with the Average Normalized Levenshtein Similarity (ANLS) score.

\textbf{Baselines.}
We compare our method with: (1) text-only pre-trained models BERT \cite{devlinBERTPretrainingDeep2019} and UniLMv2 \cite{baoUniLMv22020}; (2) a layout-aware language model BROS \cite{hongBROSPreTrainedLanguage2022}; (3) LayoutLM \cite{xuLayoutLMPretrainingText2020}, LayoutLMv2 \cite{xuLayoutLMv2MultimodalPretraining2021a}, and DocFormer \cite{appalarajuDocFormerEndtoEndTransformer2021a}, which are layout-aware Transformers with fine-grained text and visual information. (4) StructuralLM \cite{liStructuralLMStructuralPretraining2021} with coarse-grained layout information of text. (5) SelfDoc \cite{liSelfDocSelfSupervisedDocument2021} and StrucText \cite{liStrucTexT2021} with coarse-grained textual information.

\textbf{Training details.} Our model consists of \emph{Fine-grained Encoder} and \emph{Coarse-grained Encoder} and we train our models with two different parameter sizes. For the base model, we use a 12-layer 12-head Transformer encoder as the \emph{Fine-grained Encoder} and it is initialized with LayoutLMv2$_\text{base}$. For the large model, we use a 24-layer 16-head Transformer encoder initialized with LayoutLMv2$_\text{large}$. The \emph{Coarse-grained Encoder} is also a Transformer and the number of layers of it is searched from $1$ to $5$ according to the performance on validation data for different tasks.
The visual encoder in our method is also initialized with parameters of visual encoder in LayoutLMv2. We use Adam  optimizer for all tasks. The initial learning rate is 5e-5 for base model and 2e-5 for the large model. We use the weight decay of 0.01 and the learning rate is linearly warmed up and then linearly decayed. The value of radius $r$ is set to 30, 15, 30 and 50 for FUNSD, CORD, SROIE and DocVQA respectively. More details about training are provided in the supplementary materials.

\subsection{Overall Performance}

\label{sec:exp}

\begin{table}[t]
\setlength{\abovecaptionskip}{0.15cm}
\setlength{\belowcaptionskip}{0.01cm}
\caption{Performance on DocVQA}
\label{tab:exp_docvqa}
\begin{tabular}{lcc}
    \toprule
    Method &
    Information &
    ANLS \\
    \midrule
    BERT$_\text{base}$ &
    \multirow{4}{*}{Text only} &
    0.6354 \\
    BERT$_\text{large}$ &
    &
    0.6768 \\
    UniLMv2$_\text{base}$ &
    &
    0.7134 \\
    UniLMv2$_\text{large}$ &
    &
    0.7709 \\
    \midrule
    LayoutLM$_\text{base}$ &
    \multirow{4}{*}{Fine-grained text + vision + layout}&
    0.6979 \\
    LayoutLM$_\text{large}$ &
    &
    0.7259 \\
    LayoutLMv2$_\text{base}$ &
    &
    0.7808 \\
    LayoutLMv2$_\text{large}$ &
    &
    0.8348 \\
    \midrule
    mmLayout$_\text{base}$ &
    \multirow{2}{*}{Multi-grained text + vision + layout}&
    0.7915 \\
    mmLayout$_\text{large}$ &
    &
    \textbf{0.8366} \\
    \bottomrule
\end{tabular}
\end{table}

\begin{table}[t]

\setlength{\abovecaptionskip}{0.15cm}
\setlength{\belowcaptionskip}{0.01cm}
\caption{We compare the effects of different components on the performance (F1) of the model on FUNSD and CORD.}
\label{tab:exp_ablation_components}
\begin{tabular}{lcc}
    \toprule
    Model &
    FUNSD &
    CORD \\
    \midrule
    mmLayout$_\text{base}$ &
    0.8602 &
    0.9723 \\
    w/o Coarse-grained Encoder &
    0.8511 &
    0.9677 \\
    w/o Common Sense Enhancement &
    0.8574 &
    0.9631 \\
    w/o Aggregation with Cross-grained Edges &
    0.8276 &
    0.9495 \\
    \bottomrule
\end{tabular}
\end{table}

\textbf{Information Extraction tasks.}
The performance of our method and baselines on Information Extraction tasks are listed in Table~\ref{tab:exp_funsd_cord_sroie}. All layout-aware models outperform the text-only models BERT and UniLMv2. By introducing coarse-grained information, our model achieves significant improvement over LayoutLMv2. Our base model outperforms LayoutLMv2$_\text{base}$ by \textbf{3.26\%} F1 score on FUNSD, \textbf{2.28\%} F1 score on CORD, and \textbf{1.38\%} F1 score on SROIE. One possible concern is that our base model outperforms LayoutLM with the extra number of parameters of \emph{Coarse-grained Encoder}. In fact, on the contrary, our base model even outperforms LayoutLMv2$_\text{large}$ by \textbf{1.82\%} F1 score on FUNSD, \textbf{1.22\%} F1 score on CORD  while using fewer parameters. By using a larger \emph{Fine-grained Encoder}, our large model achieves better performance than our base model.

\textbf{Document Question Answering.} We evaluate the performance on DocVQA with the Average Normalized Levenshtein Similarity (ANLS) score. Our base model outperforms LayoutLMv2$_\text{base}$ by \textbf{1.07\%} ANLS score and our large model outperforms LayoutLMv2$_\text{large}$ by \textbf{0.18\%} ANLS score. Compared with the Information Extraction task, our method brings less benefit to the Document Question Answering task. One possible reason is that in our method, coarse-grained information is constructed based on the contents of the document itself, without considering the impact of the question.

\subsection{Ablation Studies}
\label{sec:exp_ablation}

\begin{table}[t]
\setlength{\abovecaptionskip}{0.15cm}
\setlength{\belowcaptionskip}{0.01cm}
\caption{Effects of the number of layers for \textit{Coarse-grained Encoder}. The ($i$,CL) represents the \emph{Coarse-grained Encoder} contains $i$ Transformer layers.}
\label{tab:ablation_layers}
\begin{tabular}{@{}ccccc@{}}
\toprule
Model& FUNSD& CORD& SROIE& \#Params\\
\midrule
LayoutLMv2$_\text{base}$& 0.8276& 0.9495& 0.9625& 200M\\
\midrule
mmLayout$_\text{base}$(1,CL)& 0.8586& \textbf{0.9723}& 0.9711& 208M\\
mmLayout$_\text{base}$(2,CL)& 0.8563& 0.9646& 0.9763& 215M\\
mmLayout$_\text{base}$(3,CL)& 0.8515& 0.9688& 0.9712& 222M\\
mmLayout$_\text{base}$(4,CL)& 0.8502& 0.9642& 0.9741& 229M\\
mmLayout$_\text{base}$(5,CL)& \textbf{0.8602}& 0.9683& 0.9733& 236M\\
\midrule
LayoutLMv2$_\text{large}$& 0.8420& 0.9601& \textbf{0.9781}& 426M\\
\bottomrule
\end{tabular}
\end{table}

\begin{table}[t]
\setlength{\abovecaptionskip}{0.15cm}
\setlength{\belowcaptionskip}{0.01cm}
\caption{Effects of radius $r$ in clustering visual regions.}
\label{tab:ablation_radius}
\begin{tabular}{@{}cccccc@{}}
\toprule
Radius & 5      & 10     & 30              & 50     & 100             \\ \midrule
FUNSD  & 0.8526 & 0.8528 & \textbf{0.8602} & 0.8509 & 0.8554 \\ \bottomrule
\end{tabular}
\end{table}

To understand the impact of components in \emph{mmLayout}, we conduct ablation studies on the base model.
\begin{figure*}[t]
  \centering
  \small
  \includegraphics[width=0.8\linewidth]{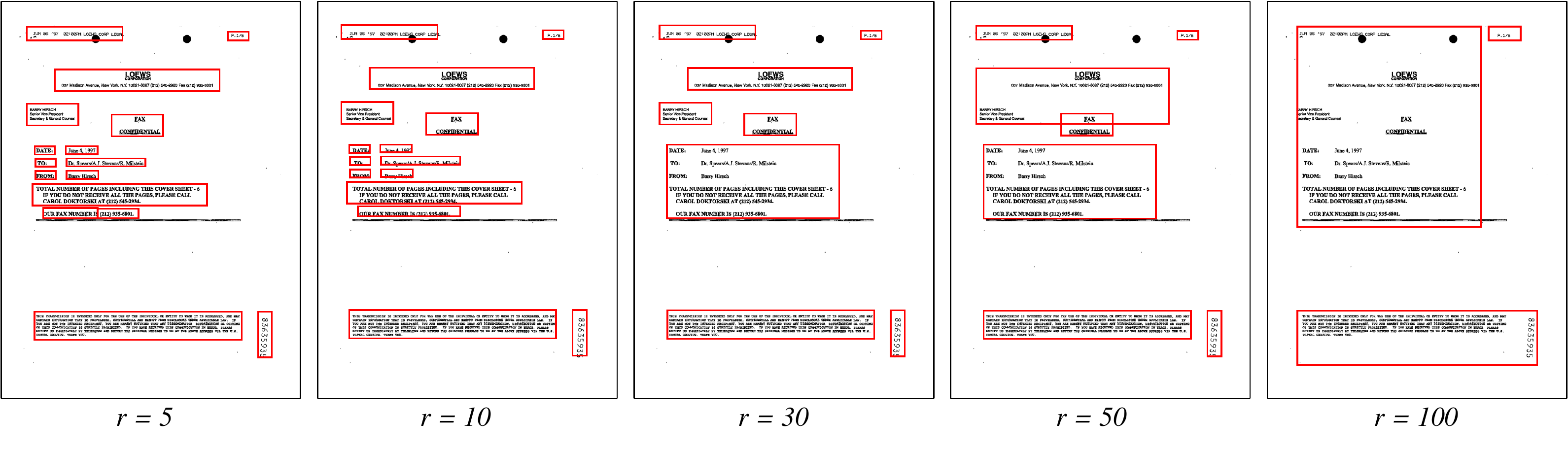}
  
  \setlength{\abovecaptionskip}{0.10cm} 
  \caption{\textbf{Visualization of salient visual regions generated by clustering with different values of radius $r$ in FUNSD.}}
  \Description{}
  \label{fig:case_study_visual_region_with_r}
\end{figure*}

\textbf{Ablation of components.}
We analyze the impact of each component on FUNSD and CORD.
Table~\ref{tab:exp_ablation_components} shows that each component is useful for document understanding. Without the aggregation with \emph{cross-grained edges}, our method degrades to LayoutLMv2 and the performance declines sharply. The Common Sense Enhancement affects CORD more than FUNSD, consistent with the fact that the entity labeling of CORD relies on more common sense.

\textbf{Ablation of layers of \emph{Coarse-grained Encoder}.} The \emph{Coarse-grained Encoder} introduces extra parameters and the impact of the number of layers of it is shown in Table~\ref{tab:ablation_layers}. We find that the performance of the model does not increase steadily with the increase of coarse-grained encoder layers. The reason is that coarse-grained and fine-grained features are fused for downstream tasks leading to competition between them, and different tasks have different degrees of preference for coarse-grained and fine-grained information. Moreover, results in Table~\ref{tab:ablation_layers} show that equipping existing base models with a few layers of \emph{Coarse-grained Encoder} can outperform existing large models, while using fewer parameters.

\textbf{Ablation of radius $r$ in clustering.} 
The number of salient visual regions detected for a document is affected by the radius $r$. When $r$ is too small, the obtained visual regions are too small to maintain high-level structural information. Conversely, a large value of $r$ results in regions that are large but not salient. Table~\ref{tab:ablation_radius} shows the affect of radius $r$ on performance of model on FUNSD. Figure~\ref{fig:case_study_visual_region_with_r} visualizes the salient visual regions corresponding to different radius. When $r=30$, the resulting visual regions are more reasonable. When $r=100$, the visual regions are too large, conversely, when $r=5$, the visual regions are too small.

\subsection{Visualization and Case Studies}

\begin{figure*}[t]
  \centering
  \small
  \includegraphics[width=0.8\linewidth]{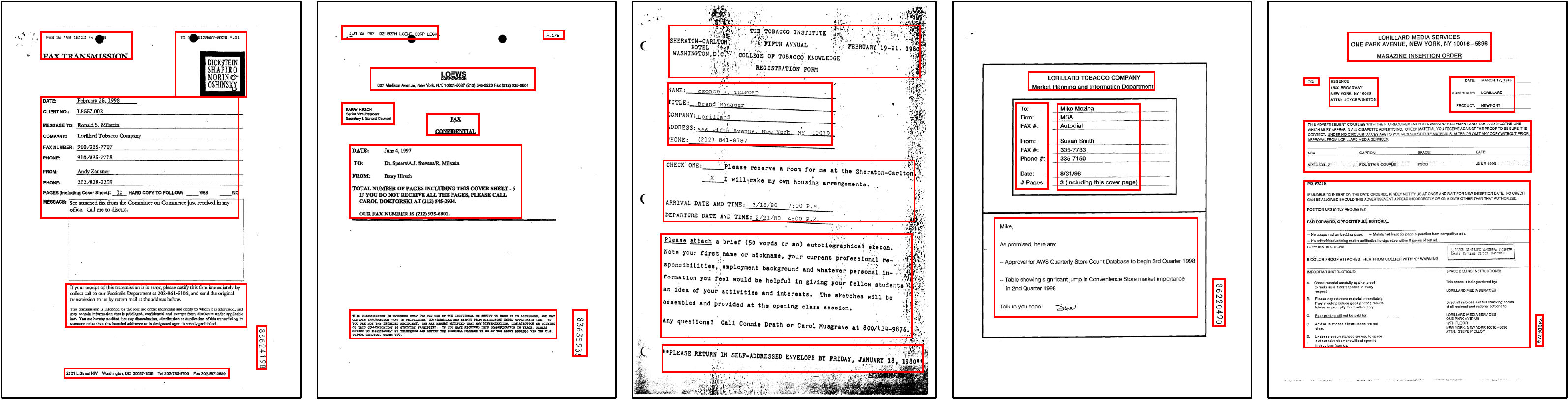}
   \setlength{\abovecaptionskip}{0.15cm} 
  \caption{\textbf{Visualization of salient visual regions of different documents generated by clustering with radius $r=30$ in FUNSD.}}
  \Description{}
  \label{fig:case_study_visual_region}
\end{figure*}

\textbf{Visualization of salient visual regions.} Figure~\ref{fig:case_study_visual_region} shows the salient visual regions of different samples detected by our method. Semantic regions, such as header regions and list regions, are detected by our method. These regions are of great help to the model in document understanding.

\begin{figure*}[t]
  \centering
  \small
  \includegraphics[width=1.0\linewidth]{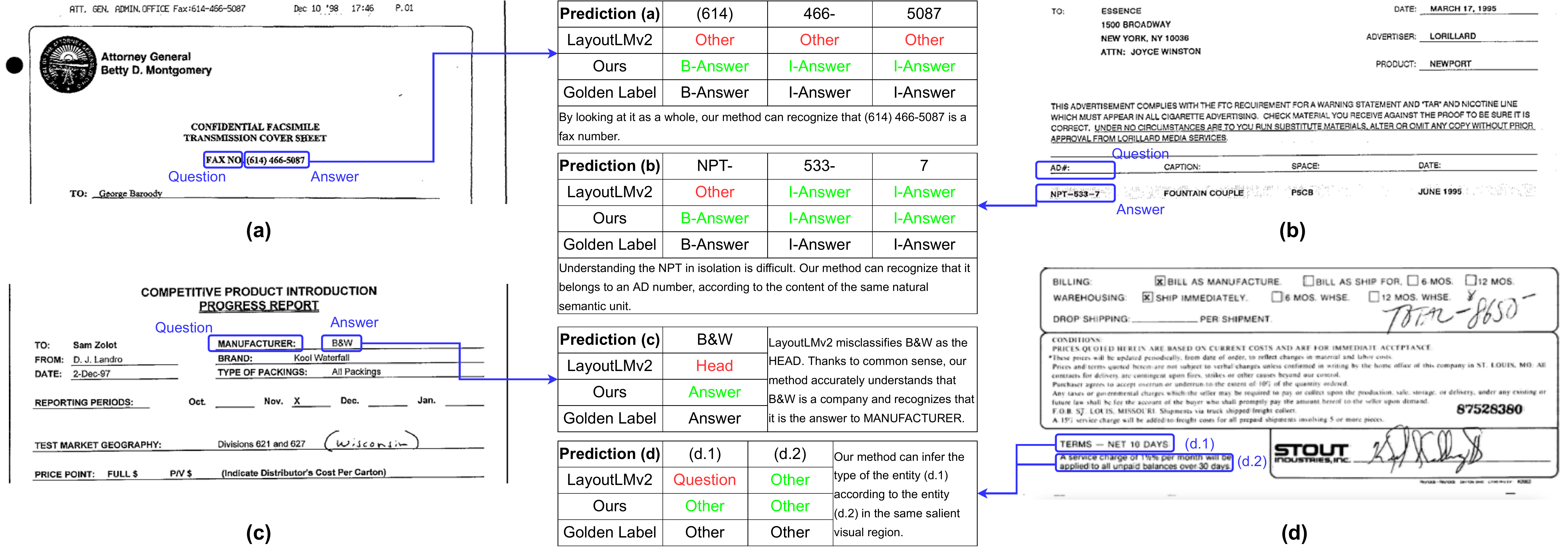}
   \setlength{\abovecaptionskip}{0.1cm}
  \caption{\textbf{Visualization of examples in FUNSD.}
  Thanks to coarse-grained information from natural lexical units and salient visual regions, as well as additional common senses, our method can recognize entities more accurately.
  }
  \Description{}
  \label{fig:case_study}
\end{figure*}

\textbf{Case studies.}
We visualize prediction results of some samples in Figure~\ref{fig:case_study}. As shown in Figure~\ref{fig:case_study}(a), it is not easy to realize that ``(614)'', ``466-'', and ``5087'' belong to a fax number by looking at them separately. By looking at them as a whole, our model can accurately recognize that they belong to the same fax number. Similarly, as shown in Figure~\ref{fig:case_study}(b), it is hard to understand the token NPT on its own, but with the help of information in the same natural semantic unit, our model realizes that it belongs to an AD number. As shown in Figure~\ref{fig:case_study}(c), thanks to the common sense enhancement, our method can accurately understand words with specific meanings, such as B\&W is a company. The sample in Figure~\ref{fig:case_study}(d) shows that our method can infer the type of an entity according to entities in the same salient visual region.

\section{Related Work}
\label{sec:related_work}
\textbf{Visually-rich document understanding.} Early works on VrDU are based on single modal or shallow multimodal fusion models. \cite{yangLearningExtractSemantic2017,kattiChargridUnderstanding2D2018,denkBERTgridContextualizedEmbedding2019,zhaoCUTIELearningUnderstand2019,sarkhelDeterministicRoutingLayout2019,zhangTRIEEndtoEndText2020,wangRobustVisualInformation2021,linViBERTgridJointlyTrained2021} exploit the layout and visual information of text in documents based on CNNs. \cite{liuGraphConvolutionMultimodal2019,qianGraphIEGraphBasedFramework2019,yuPICKProcessingKey2020,weiRobustLayoutawareIE2020,carbonellNamedEntityRecognition2021} model complex layout relationships in documents by static or dynamic graphs and extract structural information by GNNs. \cite{majumderRepresentationLearningInformation2020,wangDocStructMultimodalMethod2020} extract information from form-like document image based on language Transformers. Recently, many layout-aware pre-trained Transformers for VrDU have been proposed \cite{xuLayoutLMPretrainingText2020,appalarajuDocFormerEndtoEndTransformer2021a,garncarekLAMBERTLayoutAwareLanguage2021a,hwangSpatialDependencyParsing2021,liStructuralLMStructuralPretraining2021,liSelfDocSelfSupervisedDocument2021,liStrucTexT2021,powalskiGoingFullTILTBoogie2021,xuLayoutLMv2MultimodalPretraining2021a,xuLayoutXLMMultimodalPretraining2021,hongBROSPreTrainedLanguage2022,leeFormNetStructuralEncoding2022}. LayoutLMv2 proposes a pre-trained layout-aware multimodal Transformer based on the spatial-aware self-attention mechanism. 
TILT \cite{powalskiGoingFullTILTBoogie2021} proposes a pre-trained layout-aware multimodal encoder-decoder Transformer to unify a variety of problems involving natural language. Further, \cite{zhangEntityRelationExtraction2021,guXYLayoutLMLayoutAwareMultimodal2022b} attempt to adapt layout-aware pre-trained models to better address downstream document understanding tasks. 

However, existing layout-aware multimodal Transformers ignore the valuable coarse-grained information like natural units and salient visual regions. Several efforts take them into account but are incomplete \cite{liStructuralLMStructuralPretraining2021,liStrucTexT2021,liSelfDocSelfSupervisedDocument2021}. In this paper, we argue both fine-grained and coarse-grained multimodal information is helpful for document understanding and try to incorporate coarse-grained information into existing pre-trained layout-aware multimodal Transformers.

\textbf{Multi-grained information.} Several works have shown that multi-grained information is helpful for the understanding of both natural language and visual contents. AMBERT \cite{zhangAMBERTPretrainedLanguage2021} and LICHEE \cite{guoLICHEEImprovingLanguage2021} generate both coarse-grained and fine-grained tokens for a sentence using two different tokenizers respectively.
TNT \cite{han(TNT)TransformerTransformer2021a} divides the input images into several patches as ``visual sentence'' and then further divides them into sub-patches as ``visual words''.

\section{Conclusion}
We propose a multi-grained multimodal Transformer named \textbf{mmLayout} based on the document graph to take coarse-grained information, including natural lexical units and salient visual regions, into account for document understanding. We propose a clustering-based method to detect salient visual regions by text segments and propose a common sense enhancement strategy to exploit natural lexical units in text segments. Experimental results on Information Extraction and Document Question Answering tasks show that, with the help of coarse-grained information, our method can solve the document understanding tasks well.
In the future, we will explore how to introduce more visual information into the detection of salient visual regions, rather than only the textual layout information used in the paper.

\begin{acks}
This work was supported by the National Key R\&D Program of China (No.~2018AAA0101900), the NSFC projects (No.~62072399, No.~U19B2042, No.~61402403), Chinese Knowledge Center for Engineering Sciences and Technology, MoE Engineering Research Center of Digital Library, and the Fundamental Research Funds for the Central Universities (No.~226-2022-00070).
\end{acks}

\bibliographystyle{ACM-Reference-Format}
\bibliography{ref}


\end{document}